\documentclass{article}

\usepackage{PRIMEarxiv}

\usepackage[utf8]{inputenc} 
\usepackage[T1]{fontenc}    
\usepackage{hyperref}       
\usepackage{url}            
\usepackage{booktabs}       
\usepackage{amsfonts}       
\usepackage{nicefrac}       
\usepackage{microtype}      
\usepackage{lipsum}
\usepackage{fancyhdr}       
\usepackage{graphicx}       
\graphicspath{{media/}}     

\title{Beyond Segmentation: Road Network Generation with Multi-Modal LLMs
\thanks{\textit{\underline{Citation}}: 
\textbf{Authors. Title. Pages.... DOI:000000/11111.}} 
}

\author{
  Sumedh Rasal \\
  HERE North America LLC \\
  Georgia Institute Of Technology \\
  \texttt{srasal3@gatech.edu} \\
   \And
  Sanjay Kumar Boddhu \\
  HERE North America LLC \\
  \texttt{sanjaykumar.boddhu@here.com} \\
}

\begin{document}
\maketitle

\begin{abstract}
This paper introduces an innovative approach to road network generation through the utilization of a multi-modal Large Language Model (LLM). Our model is specifically designed to process aerial images of road layouts and produce detailed, navigable road networks within the input images. The core innovation of our system lies in the unique training methodology employed for the large language model to generate road networks as its output. This approach draws inspiration from the BLIP-2 architecture \cite{li2023blip}, leveraging pre-trained frozen image encoders and large language models to create a versatile multi-modal LLM. 

Our work also offers an alternative to the reasoning segmentation method proposed in the LISA paper \cite{lai2023lisa}. By training the large language model with our approach, the necessity for generating binary segmentation masks, as suggested in the LISA paper \cite{lai2023lisa}, is effectively eliminated. Experimental results underscore the efficacy of our multi-modal LLM in providing precise and valuable navigational guidance. This research represents a significant stride in bolstering autonomous navigation systems, especially in road network scenarios, where accurate guidance is of paramount importance.
\end{abstract}


\section{Introduction}
In recent years, the field of large language models (LLMs) has witnessed a remarkable transformation, transitioning from text-based generation to the generation of diverse modalities, including text, images, audio, and video, all through a single LLM. This evolution, from ChatGPT-4 \cite{gpt4} to a myriad of multi-modal LLMs, has significantly advanced the capabilities of AI systems to process and understand various forms of data.

These multi-modal LLMs are designed to emulate the holistic perceptual abilities of humans, enabling them to process and generate content in more versatile ways. Unlike previous models, such as ChatGPT-4 \cite{gpt4}, MiniGPT-4 \cite{zhu2023minigpt}, LISA \cite{lai2023lisa}, and others \cite{radford2021learning}, which aimed to be general-purpose multi-modal models \cite{chung2022scaling} \cite{dai2022enabling}, our work introduces a novel approach that tailors the training of such models to address a specific challenge: generating navigable road networks from aerial images.

In the groundbreaking LISA paper \cite{lai2023lisa}, the authors introduced a novel concept of utilizing both text and images within a large language model, pioneering the use of segmentation masks. While innovative, this concept prompted us to question whether segmentation masks are indispensable for this task. Could we achieve similar outcomes by training a large language model differently, while adhering to the general multi-modal architecture principles highlighted in models like MiniGPT-4 \cite{zhu2023minigpt}, NextGPT \cite{wu2023next}, BLIP-2 \cite{li2023blip}, and others?

Our model, known as NavGPT, is built upon the architecture of MiniGPT-4, harnessing Vicuna's visual component, as introduced in BLIP-2 \cite{li2023blip}. NavGPT reuses the crucial modification of a new projection layer that aligns encoded visual features with Vicuna's language model \cite{chiang2023vicuna} while keeping all other vision and language components frozen \cite{zhang2022opt} \cite{zhang2021vinvl} \cite{zhai2022lit}, introduced in MiniGPT-4's \cite{zhu2023minigpt} architecture. During training, we provide the model with a JSON file listing the image name and the precise coordinates of all road networks found in the image. The Q-Former module is trained to bridge its output with the frozen large language model (Vicuna).

This innovative training approach not only allows researchers to address unique problems using open-sourced pre-trained large language models but also serves as a testament to the versatility of large language models. In this paper, we highlight one such use case. As we transition this system into production, we aim to share insights into the challenges we've encountered, thereby demonstrating the validity of our novel training technique for mastering a wide array of tasks using an LLM. Training our model required just one A100 GPU for approximately 26 hours, highlighting the cost-effectiveness of retraining such models to cater to personalized use cases. Our objective is to showcase that large language models offer a novel solution to addressing the challenges of map-making, potentially paving the way for achieving an autonomous navigable world.

In essence, our paper makes the following pivotal contributions:
\begin{itemize}
\item \emph{Reconstructing Perception}: Leveraging the potential of large language models, we propose a novel image-instruction pair that includes the image's identifier and precise coordinates of the road network(s). This pairing empowers the large language model to develop an intrinsic comprehension of road network identification when confronted with aerial view images.
\item \emph{Segmentation Simplified}: Our unique training approach obviates the need for segmentation masks in large language models. By dispensing with the step of generating training data for region-of-interest segmentation and architecturally eschewing the production of segmented masks as model outputs, we streamline the training process and enhance efficiency.
\item Our model builds upon the robust architecture of MiniGPT-4 \cite{zhu2023minigpt}. However, it distinguishes itself by requiring a mere 10,000 image-instruction pairs for training. Remarkably, even in a zero-shot setting, our model demonstrates commendable performance, underscoring its efficiency in relation to the retraining effort. This demonstrates that upcoming multi-modal large language models can effectively address a wide array of challenges, many of which may not be solely text-based.
\end{itemize}

In summary, our research builds upon the recent advancements in multi-modal LLMs \cite{lai2023lisa} \cite{li2022blip} \cite{li2023blip} \cite{wu2023next} \cite{tsimpoukelli2021multimodal} \cite{zhu2023minigpt} \cite{koh2023generating}, providing a focused and innovative solution to the task of generating navigable road networks from aerial images. By leveraging a tailored training approach, NavGPT aims to empower autonomous navigation systems, particularly in scenarios where precise navigational guidance is essential.

\section{Related Works}

\subsection{Evolution of Large Language Models}
The quest for progress in Artificial General Intelligence (AGI) has been an enduring aspiration within the research community, with numerous tools and methodologies explored \cite{dong2019unified} \cite{zhu2023chatgpt}. However, a true breakthrough remained elusive. Everything changed with the introduction of GPT-3, particularly the emergence of ChatGPT \cite{chatgpt}, which harnessed the power of GPT-3. The journey of GPT-based models has been a remarkable evolution, marked by continuous advancements. The most recent stride, embodied in ChatGPT powered by GPT-4  \cite{gpt4}, has redefined the landscape of general artificial intelligence.

Yet, the inner workings of GPT-4 \cite{gpt4} remained shrouded in mystery for the broader research community. This enigma persisted until the advent of open-sourced models such as Vicuna \cite{chiang2023vicuna} and LLaMa \cite{touvron2023llama} \cite{wei2022emergent} \cite{taori2023stanford}. Each of these models brought noteworthy enhancements in terms of retraining possibilities and the quality of inferred model outputs. Notably, Meta's recent release of LLaMa-2 \cite{touvron2023llama2} designed for commercial applications, represents a significant development. This newfound accessibility is expected to foster innovation across various domains.

\subsection{Foundational Multi-Modal Large Language Models}
As Large Language Models (LLMs) \cite{chiang2023vicuna} \cite{devlin2018bert} \cite{taori2023stanford} \cite{touvron2023llama} \cite{touvron2023llama2} \cite{zhu2023chatgpt} \cite{fang2023eva} \cite{goyal2017making} \cite{guo2022images} \cite{cho2021unifying} \cite{chen2022pali} \cite{yang2023mm} \cite{yang2022zero} \cite{driess2023palm} \cite{huang2023language} began to conquer text generation challenges, a world of new possibilities unfurled. Notably, they exhibited an improved grasp of contextual nuances, leading to more coherent text-to-text conversations. Extensive efforts were dedicated to incorporating human-in-the-loop feedback, refining the conversational finesse of LLMs. However, it was inevitable that the research landscape would expand beyond text-to-text generation alone.

Soon enough, the research community shifted its focus to text-and-image conversations as inputs for LLMs, heralding the era of Multi-Modal Large Language Models. The fundamental idea driving this approach involved training the models to comprehend images by processing visual features through dedicated encoders. These features were then seamlessly integrated into LLMs as input. This methodology greatly facilitated the LLMs' capacity to interpret images, harnessing the valuable information embedded in image-caption pairs during the training phase.

\subsection{Advancing Multi-Modal Large Language Models}
In recent times, the latest iterations of multi-modal large language models \cite{lai2023lisa} \cite{li2022blip} \cite{li2023blip} \cite{tsimpoukelli2021multimodal} \cite{wu2023next} \cite{zhu2023minigpt} have embraced versatility, accommodating an array of input formats. These formats encompass text, images, videos, and audio, reflecting the dynamic nature of contemporary data sources  \cite{koh2023generating}. Correspondingly, these models exhibit a harmonious synergy between inputs and outputs. For each distinct format, a dedicated encoder and decoder are seamlessly integrated. This structural architecture empowers advanced systems like ChatGPT-4 \cite{gpt4} to seamlessly process input data and generate corresponding outputs.

Nevertheless, even these advanced systems are not immune to imperfections, particularly within their encoders and decoders. These systems may encounter challenges related to information loss during data transformation, leading to instances where the broader contextual understanding is compromised. This paper introduces novel solutions to address a few of these challenges.

\section{Method}
While multi-modal large language models offer numerous advantages, a significant hurdle lies in the generation of training data. Our model is rooted in the architectural framework of MiniGPT-4 \cite{zhu2023minigpt}, designed to establish synergy between visual data from a pre-trained vision encoder and a sophisticated large language model (LLM). We leverage Vicuna \cite{chiang2023vicuna} as the language decoder, introducing a pioneering training method for road network identification within images.

In terms of visual perception, we adopt a similar approach to the visual encoder employed in BLIP-2 \cite{li2023blip}, harnessing a Vision Transformer \cite{dosovitskiy2020image} backbone in conjunction with their pre-trained Q-Former \cite{li2023blip}. To facilitate seamless communication between the visual encoder and the LLM, we introduce a novel training procedure, augmenting the MiniGPT-4 \cite{zhu2023minigpt} projection layer. This innovative approach bridges the gap, enabling effective collaboration between the two components.

\begin{figure}
  \centering
  \includegraphics[width=.9\columnwidth]{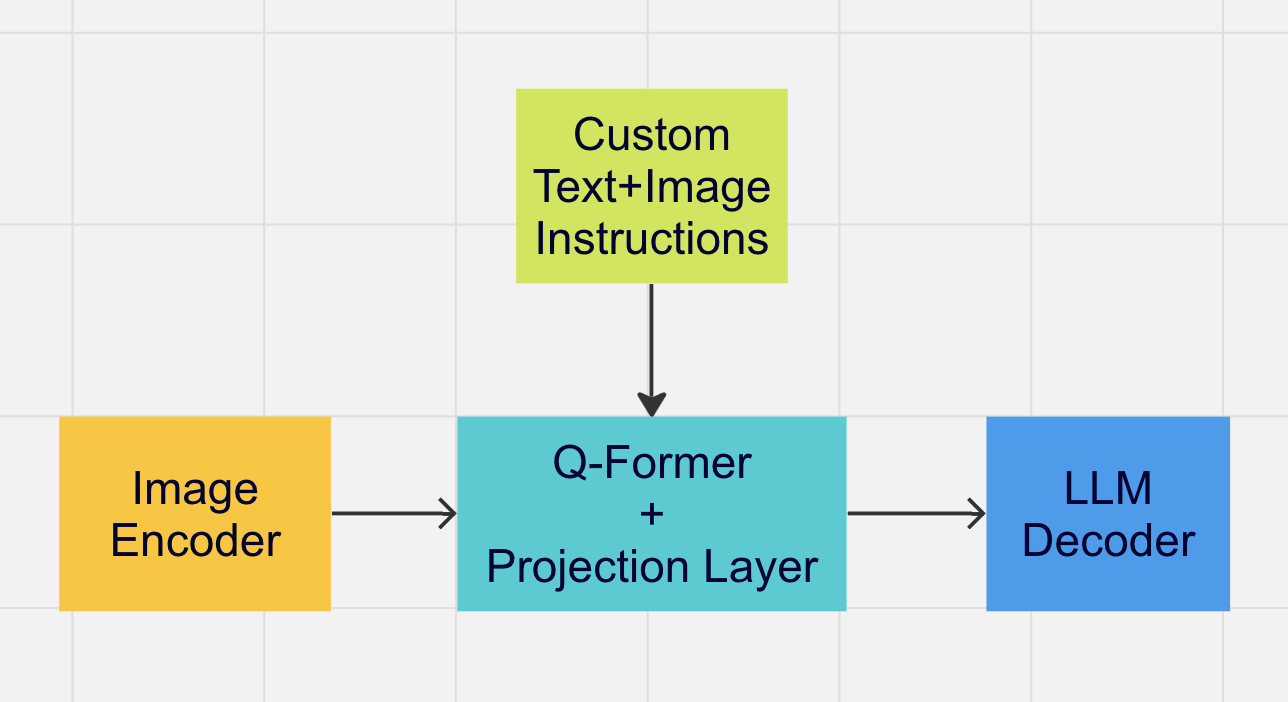}
  \caption{NavGPT architecture based on MiniGPT-4 Source: \cite{zhu2023minigpt}.}
  \label{fig:fig1}
\end{figure}

\subsection{Data Collection and Preprocessing}
Automating road navigation presents a formidable challenge, primarily due to the labor-intensive nature of data collection. However, our work benefits from operating within the spatial domain, where obtaining a portion of the required training data is relatively straightforward. We are privileged to have access to satellite imagery for specific global regions. In this study, we focus on training the model using imagery sourced from the Western European region.

To infuse a higher level of naturalness into the generated language and elevate the model's overall utility, we advocate for the importance of road navigation segment training. Unfortunately, datasets suitable for the vision-language domain, especially in the context of road navigation, are virtually non-existent. To overcome this gap, we painstakingly assembled a meticulously detailed image description dataset. This dataset has been meticulously crafted with the sole purpose of facilitating the alignment of vision and language. During the alignment phase of our NavGPT model, this dataset is deployed to fine-tune the model for enhanced performance.

\subsection{Novel Training Approach}
NavGPT, our innovative approach, deviates from the reliance on segmentation masks suggested by LISA \cite{lai2023lisa}. Our model draws inspiration from MiniGPT-4 \cite{zhu2023minigpt}, which is itself influenced by BLIP-2 
\cite{li2022blip} \cite{li2023blip}. The pivotal component, the projection layer coupled with the Q-Transformer \cite{li2023blip}, empowers our model to assess the presence or absence of a road network in a given image.

This accomplishment is realized through training the model with an image-instruction pair file that explicitly indicates whether a given image contains a road network. Here is a snippet of the JSON file contents, which sheds light on our training data.

\begin{verbatim}
    {
        "image_id": "54534_33840",
        "caption": "Found a road"
    },
    {
        "image_id": "54537_33868",
        "caption": "No roads found"
    }
\end{verbatim}

We conducted model training for 5,000 steps and observed promising results in a zero-shot setting. This initial success ignited our curiosity about the model's potential. What if, in addition to recognizing the presence of a road network within an image, we could train it to pinpoint the image coordinates of the road network? To explore this hypothesis, we required a substantial dataset.

Fortunately, we had access to various road geometries in the Western European region. Leveraging HERE's internal aerial imagery service, we initiated image queries in regions where road network geometry overlapped. The images had dimensions of 1280 by 1280 pixels. These overlapping regions corresponded to areas with well-defined road networks. A Python function was employed to perform these queries and extract the image coordinates of the road network (in the form of a line string) for 10,000 such scenarios.

Notably, approximately 40\% of the images did not contain overlapping road network geometry. However, these images were still included in the original set of 10,000 instructions. This inclusion served the dual purpose of enabling the model to discern the presence or absence of a road network and, if present, to provide accurate image coordinates. 

\begin{verbatim}
    {
        "image_id": "54537_33867",
        "caption": Found 1 road. Image coordinates are as follows: 
        [[(219, 114), (283, 271)]]"
    },
    {
        "image_id": "54537_33879",
        "caption": "Found 2 roads. Image coordinates are as follows: 
        [[(0, 775), (0, 731), (644, 28)], [(365, 0), (629, 3), (644, 28)]]"
    }
\end{verbatim}

\section{Experiment}
\begin{figure}
  \centering
  \includegraphics[width=.9\columnwidth]{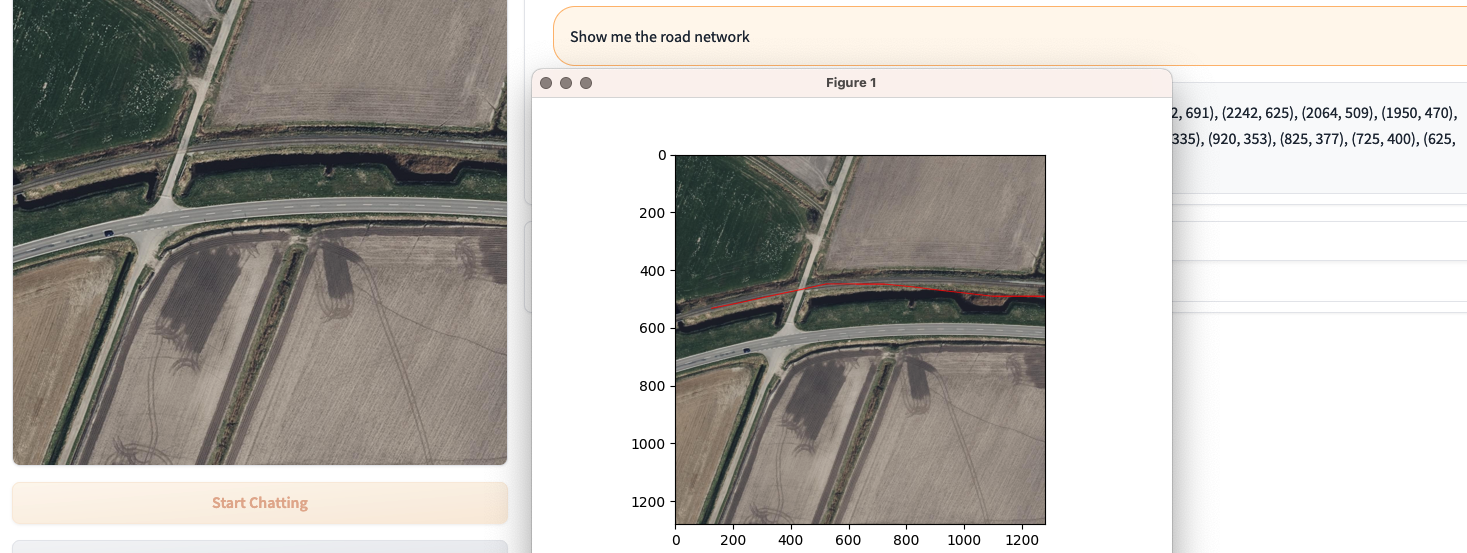}
  \caption{Experiment 1: Identify road network}
  \label{fig:fig2}
\end{figure}

\begin{figure}
  \centering
  \includegraphics[width=.9\columnwidth]{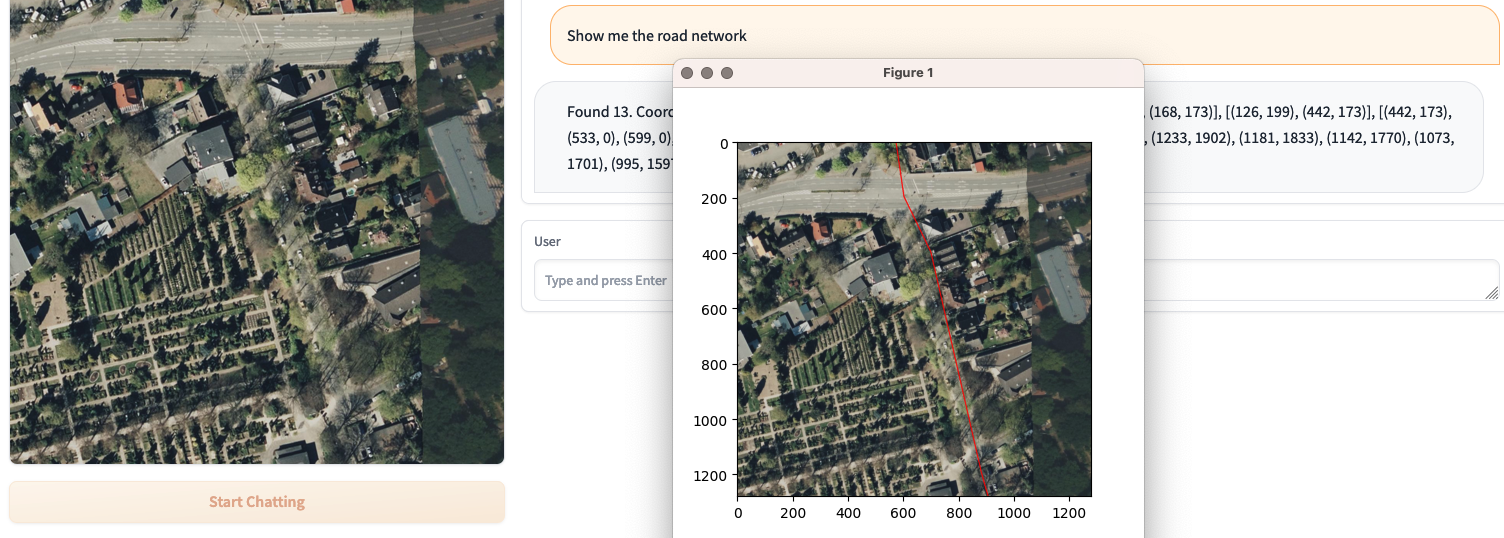}
  \caption{Experiment 2: Identify road network}
  \label{fig:fig3}
\end{figure}

In this section, we embark on a journey to unravel the diverse and emerging capabilities of our NavGPT model. Through a series of qualitative experiments, we shed light on NavGPT's remarkable proficiency in a spectrum of navigation-based tasks, showcasing its advanced abilities compared to traditional vision-language models. Refer figures \ref{fig:fig2} \ref{fig:fig3}

To assess our model, we implemented a straightforward method to compare its output with the ground truth data acquired during the training phase. We reserved a set of 100 image-instruction pairs for the testing phase.

Table 1 \ref{tab:table1} presents the model's accuracy in discerning the presence or absence of a road network in an image, with a score of 0.69.

In Table 2 \ref{tab:table2}, we focus on the model's ability to identify the number of roads within a given image, achieving an accuracy score of 0.37. We believe that further refinement can be achieved by extending the model's training to 2,000 - 5,000 additional steps.

\begin{table}
 \caption{Qualitative Evaluation 1: Confusion Matrix}
  \centering
  \begin{tabular}{lll}
    \toprule
    \multicolumn{2}{r}{NavGPT}                   \\
    \cmidrule(r){2-3}
         &  Road Found    & Road Not Found \\
    \midrule
    Ground Truth Road Found         & 61 & 13 \\
    GT Road Not Found               & 18 & 8  \\
    \bottomrule
  \end{tabular}
  \label{tab:table1}
\end{table}

\begin{table}
 \caption{Qualitative Evaluation 2}
  \centering
  \begin{tabular}{lll}
    \toprule
    \multicolumn{2}{r}{NavGPT}                   \\
    \cmidrule(r){2-3}
         &  Roads with 1 Road    & Roads with 2+ Roads \\
    \midrule
    GT 1 Road Found         & 12 & 3 \\
    GT 2+ Roads Found       & 19 & 1  \\
    \bottomrule
  \end{tabular}
  \label{tab:table2}
\end{table}

\subsection{Road Navigation Interpretation}
NavGPT's exceptional capability lies in its proficiency to offer detailed and contextually relevant road networks based on the input image. To exemplify this, we conducted a performance comparison between NavGPT and MiniGPT-4 \cite{zhu2023minigpt}, the foundational model upon which NavGPT is built. Through two distinct examples, we unveil NavGPT's multifaceted descriptive abilities. One example from our evaluation highlights NavGPT's adeptness at recognizing multiple road networks within an image. In sharp contrast, MiniGPT-4 \cite{zhu2023minigpt} lacks the training to discern road networks, underscoring the evident advantage of NavGPT's nuanced comprehension of visual content. Unlike previous iterations of multi-modal large language models, which were crafted to be general-purpose, NavGPT steps beyond this constraint to address real-world challenges in a relevant context.

\subsection{Limitations}
Given the NavGPT model training approach, we are moving away from a general-purpose multi-modal language \cite{lai2023lisa} \cite{li2022blip} \cite{li2023blip} \cite{tsimpoukelli2021multimodal} \cite{wu2023next} \cite{zhu2023minigpt} to make it a navigation-based model which means a small amount strips away the general-purpose utility of this model. The context of our training data improves the whole representation of the model to solve/understand one of the most critical problems we face from an autonomous world perspective. 

The model exhibits a reasonable performance, although it does encounter some limitations. In our forthcoming research within this domain, we intend to expand the range of scenarios presented in the model and undertake retraining to enhance its overall accuracy.

\section{Conclusion}
In conclusion, NavGPT presents a novel way to train the multi-modal large language models. By diverging from traditional segmentation masks and leveraging the architecture of MiniGPT-4 \cite{zhu2023minigpt} influenced by BLIP-2 \cite{li2022blip} \cite{li2023blip}, NavGPT empowers itself to discern and describe road networks in images. Our approach, founded on the projection layer and the Q-Transformer \cite{li2023blip}, offers a nuanced understanding of visual content, surpassing the capabilities of its predecessors.

Through extensive training with image-instruction pairs, NavGPT has demonstrated impressive abilities in identifying and describing road networks within images. The model's success, even in zero-shot settings, has inspired further exploration. By introducing image coordinates into the training process, we aim to unlock even greater potential.

Our work showcases the advancements in multi-modal large language models and addresses a real-world problem under the right context. NavGPT's contribution extends beyond the scientific community, offering practical applications for developing autonomous navigation systems.

In summary, NavGPT represents a remarkable leap in multi-modal AI, heralding a new era in understanding and generating diverse modalities from text and images, with the potential to transform the field of AI in ways we are only beginning to fathom.

\section*{Acknowledgments}
We extend our gratitude to HERE North America LLC for generously providing the hardware necessary for model training and conducting our experiments. We also appreciate HERE for granting us access to their aerial imagery service and the road network line strings that were instrumental in NavGPT's training.

\bibliographystyle{unsrt}  
\bibliography{references}  

\end{document}